\documentclass[letterpaper]{article} 
\usepackage{aaai25}  
\usepackage{times}  
\usepackage{helvet}  
\usepackage{courier}  
\usepackage[hyphens]{url}  
\usepackage{graphicx} 
\usepackage{natbib}  
\usepackage{caption} 
\usepackage{algorithm}
\usepackage{algorithmic}

\usepackage{subcaption}
\usepackage{pgfplots}
\usepgfplotslibrary{groupplots}
\usetikzlibrary{patterns}
\pgfplotsset{compat=1.18}
\usepackage{xcolor}
\usepackage{amsmath, amssymb, amsthm, mathtools}
\usepackage{booktabs}
\usepackage{lineno}
\usepackage{cleveref}
\usepackage{tikz}
\usetikzlibrary{automata, positioning}

\usepackage{diagbox}

\usepackage{multirow}
\usepackage{amsmath}
\usepackage{amssymb}
\usepackage{makecell}

\usepackage{newfloat}
\usepackage{listings}
\DeclareCaptionStyle{ruled}{labelfont=normalfont,labelsep=colon,strut=off} 
\floatstyle{ruled}
\newfloat{listing}{tb}{lst}{}
\floatname{listing}{Listing}
\title{JoyType: A Robust Design for Multilingual Visual Text Creation}
\author{
    Chao Li\textsuperscript{\rm 1},
    Chen Jiang\textsuperscript{\rm 2},
    Xiaolong Liu\textsuperscript{\rm 1},
    Jun Zhao\textsuperscript{\rm 1},
    Guoxin Wang\textsuperscript{\rm 1}
}
\affiliations{
    \textsuperscript{\rm 1}JD Health International Inc., China\\
    \textsuperscript{\rm 2}University of Chinese Academy of Sciences, China\\
    chaolee.xd@gmail.com, jiangchen22@mails.ucas.ac.cn, \{zhaojun10, liuxiaolong10, wangguoxin14\}@jd.com
}

\begin{document}

\maketitle
\footnotetext{This paper is currently under review at AAAI 2025.}

\begin{abstract}

Generating images with accurately represented text, especially in non-Latin languages, poses a significant challenge for diffusion models. Existing approaches, such as the integration of hint condition diagrams via auxiliary networks (e.g., ControlNet), have made strides towards addressing this issue. However, diffusion models often fall short in tasks requiring controlled text generation, such as specifying particular fonts or producing text in small fonts.
In this paper, we introduce a novel approach for multilingual visual text creation, named JoyType, designed to maintain the font style of text during the image generation process. Our methodology begins with assembling a training dataset, JoyType-1M, comprising 1 million pairs of data. Each pair includes an image, its description, and glyph instructions corresponding to the font style within the image. We then developed a text control network, Font ControlNet, tasked with extracting font style information to steer the image generation.
To further enhance our model's ability to maintain font style, notably in generating small-font text, we incorporated a multi-layer OCR-aware loss into the diffusion process. This enhancement allows JoyType to direct text rendering using low-level descriptors.
Our evaluations, based on both visual and accuracy metrics, demonstrate that JoyType significantly outperforms existing state-of-the-art methods. Additionally, JoyType can function as a plugin, facilitating the creation of varied image styles in conjunction with other stable diffusion models on HuggingFace and CivitAI. Our project is open-sourced on https://jdh-algo.github.io/JoyType/.

\end{abstract}

\section{Introduction}
\label{sec::Introduction}

\begin{figure}[t]
    \centering
        \includegraphics[width=0.90\columnwidth]{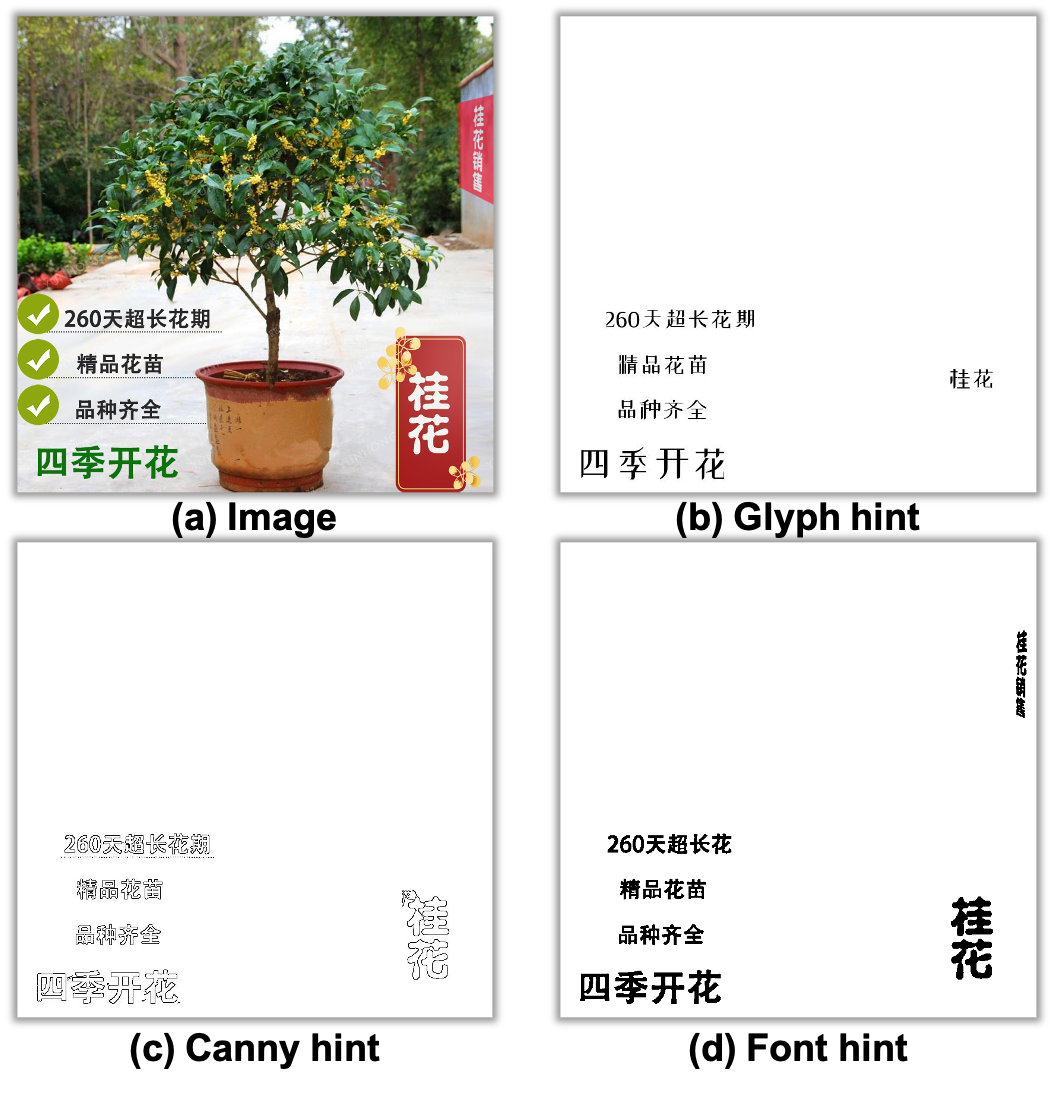}
    \caption{Compared to the commonly used glyph hint (b), JoyType introduces two new kinds of hint instructions: (c) Canny hint and (d) Font hint. }
\label{fig::hint}
\end{figure}

\begin{figure*}[t]
    \centering
        \includegraphics[width=0.98\textwidth]{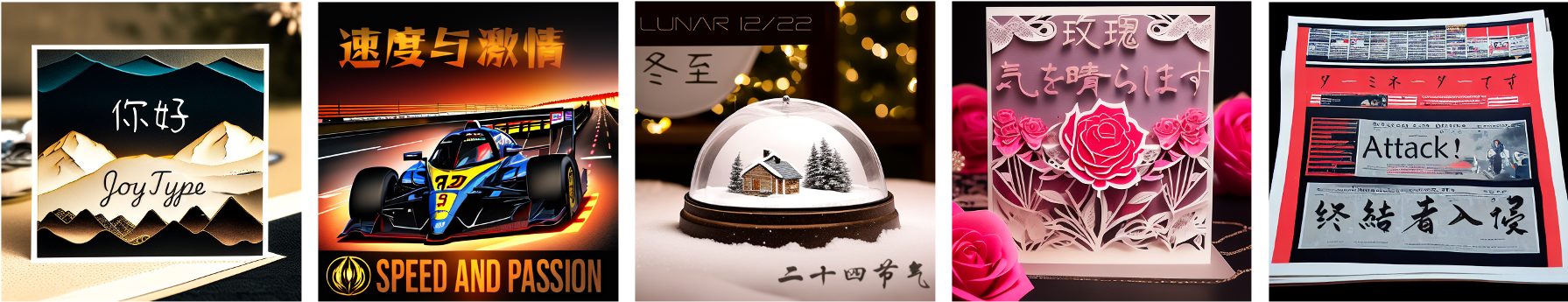} 
    \caption{Illustration of JoyType's capacity to render high-fidelity multilingual text images.}
    \label{fig::samples}
    \vspace{-0.25cm}
\end{figure*}

The success of the stable diffusion model has significantly enhanced the quality of image generation. Building upon this foundational model, Zhang et al.~\cite{zhang2023adding} proposed ControlNet, which introduces specific control conditions (such as Canny, Depth, etc.) to enable the diffusion model to perform designated generation tasks. Additionally, Hu et al.~\cite{hu2021lora} introduced the LoRA architecture, which integrates an extremely lightweight model structure into the foundational model to control the style of the image or elements within the image. These advancements have brought the diffusion model significantly closer to practical applications.

Recent research has increasingly focused on the field of visual text rendering, a challenging task that comprises two main aspects. On one hand, the model needs to accurately understand the prompt from users, including grasping the semantics and distinguishing between the scene to be generated and the text content. On the other hand, it needs to accurately render the text into the image. To address these challenges, preliminary attempts have been made, and these methods can be broadly categorized into two technical pathways.
The first type involves designing a new text encoder rather than directly adopting open-source models (e.g., CLIP). This new text encoder converts text characteristics, such as font style, and color, into tokens. By training their designed text encoder, the model can recognize the target text to be rendered in the prompt. However, the drawback of this approach is quite evident: the range of text to be generated must be specified before their model training, making the well-trained model cannot handle the text never seen before.
The second type of method involves designing a control network to guide the foundational diffusion model, assisting it in completing the text generation task. Previous methods uniformly used glyphs as hint conditions, aiming for the control network to learn glyph information and assist the foundational model in learning text rendering through cross-attention mechanisms. Because they used a generic font style as glyph instruction (e.g., Arial Unicode), there is a significant gap between this uniform glyph and the font styles in the original images. Consequently, it is necessary to collect a large amount of training data to enable the control network to bridge this gap. This approach endows visual text rendering with diversity. However, it is a method where glyphs are uncontrollable, meaning that the glyphs cannot be maintained during text generation, making it unsuitable for applications requiring precise control, such as in the design domain.
To control font styles in visual text rendering, a feasible approach is to use hint instructions that visually represent font styles. As shown in Figure~\ref{fig::hint}, this paper introduces two new hint instructions: Canny and Font hint. The Canny hint is obtained by extracting edge around the text in the original image, while the Font hint is obtained by segmenting the text in the original image. Compared to the Glyph hint, these hints are closer to the text style in the reference image. Therefore, a model trained using these two hints can generate images with font styles that are closer to those in the reference image.

In this paper, we present a robust design for multilingual visual text rendering, called JoyType. Specifically, JoyType offers a novel approach to visual text rendering by incorporating a Font ControlNet, which enables accurate text rendering and font style control. Additionally, we have developed a new loss function to supplement the latent diffusion loss, termed the multi-layer OCR perceptual loss, aimed at improving the quality of small font generation. We highlight the contributions of this work as follows:
\begin{itemize}
    \item For the task of multilingual visual text rendering, we offer a novel solution: JoyType. JoyType employs font hint conditions for text glyph instructions, enabling the control network to provide more precise guidance and thereby reducing training complexity.
    \item To fully leverage the perceptual capabilities of deep convolutional networks for low-level image descriptors, we have designed a new multi-layer OCR perceptual loss. This enhancement significantly improves the model's capability in rendering small-sized text.
    \item We evaluate the proposed JoyType by rendering multilingual text across various languages and multiple font styles. Extensive results demonstrate the effectiveness of the multi-layer OCR perceptual loss in JoyType.
\end{itemize} 
\section{Related Work}

\begin{figure*}[t]
\centering
    \includegraphics[width=0.98\textwidth]{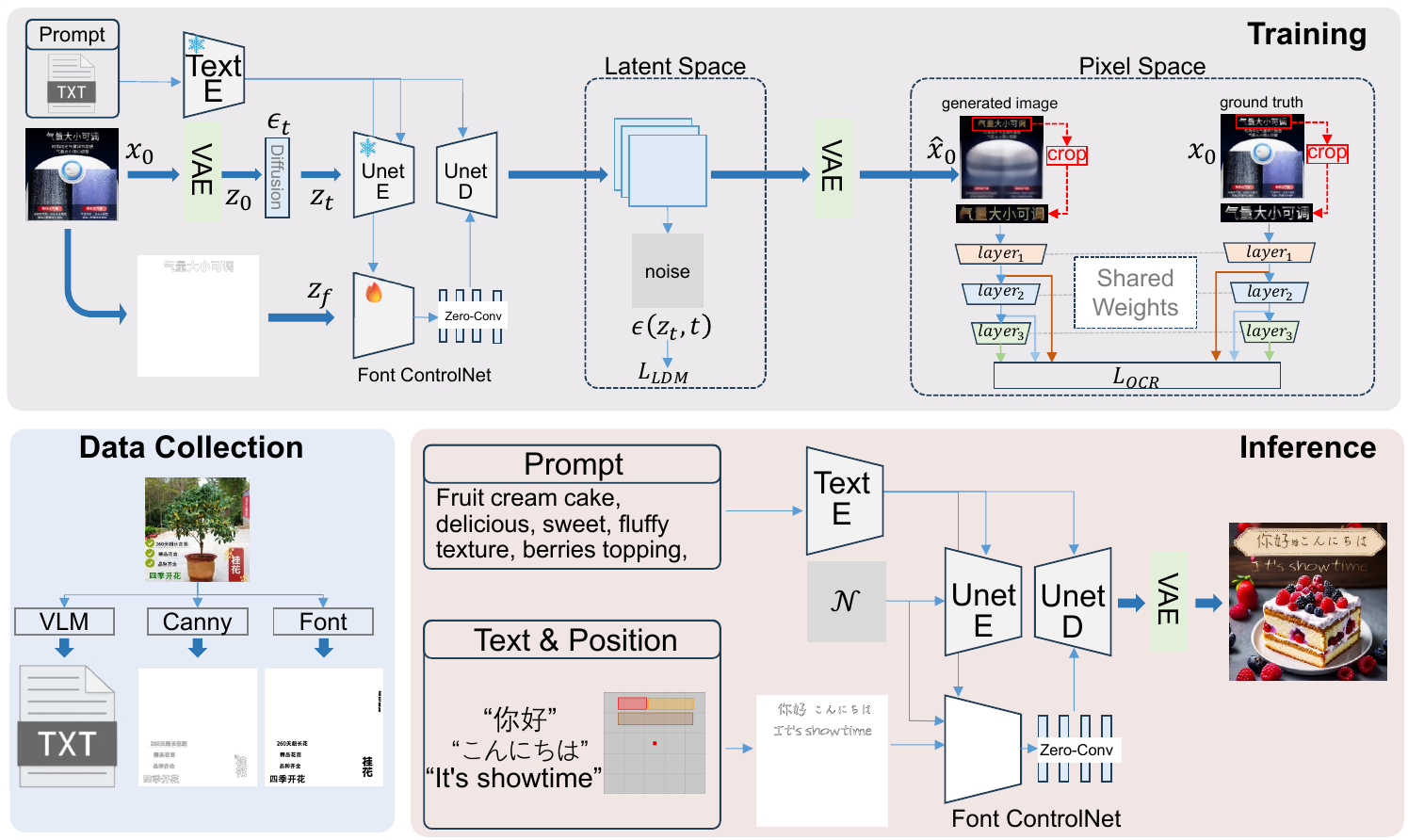}
\caption{The comprehensive framework of JoyType, illustrating the training pipeline, inference process, and data collection.}
\label{fig::framework}
\vspace{-0.25cm}
\end{figure*}

\textbf{Text-to-image Diffusion Models.} 
Denoising Diffusion Probabilistic Model~\cite{ho2020denoising} demonstrates impressive image generation capabilities, the subsequent work~\cite{ramesh2022hierarchical, rombach2022high, saharia2022photorealistic} also demonstrated the possibility of using text prompts for high-quality image generation. 
GLIDE~\cite{nichol2021glide} emphasizes the necessity of the classifier-free guidance over CLIP~\cite{radford2021learning} guidance in high-resolution generation. 
The appearance of Latent Diffusion Model~\cite{rombach2022high} successfully put the diffusion process of the image into the latent space, which greatly reduces the raining and inference costs. 
Stable Diffusion is an application of the Latent Diffusion Model, shows remarkable text-to-image generation ability by training on larger datasets. 
SDXL~\cite{podell2023sdxl} uses the U-Net with larger parameters, while introducing new refinement strategies to further improve the quality of generated images. 
Unlike the aforementioned U-Net based diffusion models, Stable Diffusion3~\cite{esser2024scaling} uses the architecture in DiT~\cite{peebles2022scalable}, and obtains more semantic information by concatenating the text embeddings from CLIP-G, CLIP-L and T5~\cite{raffel2020exploring}, thus demonstrating the further capacity in image generation. 
In ours work, we select Stable Diffusion as the base model.

\textbf{Controllable Image Generation.} 
To achieve more controlled generation of diverse content, the segmentation maps or depth maps could be input into the Diffusion Model~\cite{rombach2022high}. 
Beyond this intuitive strategy, other diffusion-based image editing techniques~\cite{meng2021sdedit, kawar2023imagic, gal2022image}, show promise in managing the content of synthetic images.
Composer~\cite{huang2023composer} decomposes the image synthesis process into several factors and then recombines them to generate new images.
Both T2IAdapter~\cite{mou2024t2i} and ControlNet~\cite{zhang2023adding} introduce a new network bypass to incorporate additional image information such as edge and depth, demonstrating the ability to accurately control object structure and color without affecting the performance of the original model.
With the appearance of IP-Adapter~\cite{ye2023ip}, multiple images can be used as the image prompt simultaneously, which greatly improves the consistency between the generated image and the original image.

\textbf{Visual Text Rendering.} Text rendering is a critical task in controllable image generation, aiming to generate accurate and well-laid-out text on images while seamlessly blending with the background. 
Imagen~\cite{saharia2022photorealistic}, eDiff-I~\cite{balaji2022eDiff-I}, and DeepFloyd IF~\cite{deepfloyd} leverage large-scale language models to enhance text spelling knowledge and train character-aware variants to address the issue of encoder insensitivity to token length. These methods, however, still face challenges in accurately rendering text.
The TextDiffuser series~\cite{chen2023textdiffuser1} and~\cite{chen2023textdiffuser2} employ layout transformers and large language models (LLMs) to predict the layout of input prompts, achieving layout automation. Despite these advancements, the accuracy of the generated text remains a challenge, and these methods do not support the generation of non-Latin scripts.
UDiffText~\cite{zhao2023udifftext} and Glyph-ByT5~\cite{liu2024glyph} design and train text encoders that are character-aware and glyph-aligned, providing more robust text embeddings as conditional guidance. However, they lack the flexibility to generate characters that were not included in the training set, limiting their extensibility.
GlyphDraw~\cite{ma2023glyphdraw} modifies the network structure to utilize glyph and positional information for drawing characters. GlyphControl~\cite{yang2024glyphcontrol} and Brush Your Text~\cite{zhang2024brush} enhance text-to-image diffusion models by leveraging glyph shape information through a ControlNet branch. Brush Your Text further introduces local attention constraints to address unreasonable text placement in scenes.
AnyText~\cite{tuo2023anytext} incorporates an auxiliary latent module and a text embedding module in its diffusion pipeline, using text-controlled diffusion loss and text-aware loss during training to enhance writing accuracy. However, it lacks the ability to maintain font styles and generate small font text.

\begin{figure*}[t]
    \centering
    \includegraphics[width=0.98\textwidth]{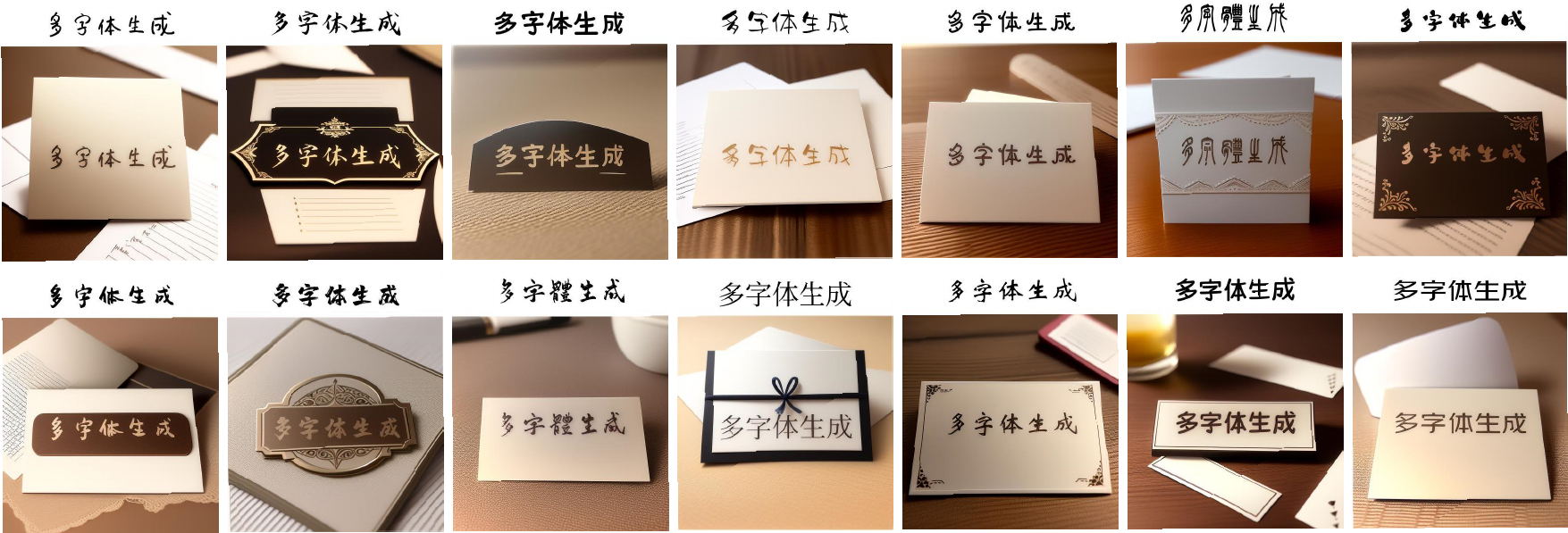} 
    \caption{Using various different font styles as hint condition images to evaluate JoyType's ability to maintain glyphs. All images use the same prompt of ``a card." We label the standard style of each font (hint image) at the top of each image.}
    \label{fig::font}
    \vspace{-0.25cm}
\end{figure*}

\section{Proposed JoyType}
\label{sec::Methodology}
As illustrated in Fig.\ref{fig::framework}, the whole framework of our JoyType companies three main parts, including the training pipeline, inference pipeline, and data collection. The overall learning objective of the entire diffusion model is bifurcated into two segments: the latent space and the pixel space. Within the latent space, we utilize the loss function $L_{ldm}$ associated with Latent Diffusion Models. The latent features are then decoded back into images via the  Variational Autoencoder (VAE)~\cite{kingma2013auto} decoder. Within the pixel space, the text regions of both the predicted and the ground truth images are cropped and processed through an OCR model independently. We extract the convolutional layer features from the OCR model and compute the Mean Squared Error (MSE) loss between the features of each layer, thereby constituting the loss $L_{ocr}$. In the following sections, we will introduce each aspect.

\subsection{Text-Control Training Pipeline}
In the training pipeline, we introduce the text-control generation, which comprises three primary components: the latent diffusion module, the Font ControlNet module, and the loss design module. Following the work of ControlNet, JoyType employs hint-guided conditioning and cross-attention mechanisms in the latent space to facilitate diffusion learning for images. More precisely, to train JoyType model, the raw image, canny or font hint instruction, and prompt are fed into the VAE, Font ControlNet, and text encoder, respectively. 
In the text-control diffusion pipeline, we first generate a latent representation $z_0 \in \mathbb{R}^{m \times n \times c} $ by applying the VAE to the input image $ x_0 \in \mathbb{R}^{M \times N \times 3} $. Here, $ m \times n $ denotes the feature resolution, and $ c $ represents the latent feature dimension. Subsequently, latent diffusion algorithms incrementally add noise to $ z_0 $, resulting in a noisy latent image $ z_t $ at each time step $t$.

Given conditions that include the time step $t$, an guidance feature $ z_f \in \mathbb{R}^{m \times n \times c} $ produced by the Font ControlNet module, and a text embedding $ c_{t} $ generated by the text encoder module, the noise added to the noisy latent image $ z_t $ can be predicted by a network $\varepsilon_{\theta}$, thus further can obtain the predict image with

\begin{equation} \label{eq::loss-ldm}
    L_{LDM} = \mathbb{E} _{z_{0}, t, c_{t}, z_{f},\varepsilon \sim \mathcal{N}\left ( 0,1 \right )   } \left [ \left \| \varepsilon - \varepsilon_{\theta }\left ( z_{t}, t, c_{t}, z_{f}  \right )   \right \| _{2}^{2}  \right ],
\end{equation}
where $L_{ldm}$ represents the objective function for finetuning Font ControlNet in the latent space under font instructions.

\subsection{Multi-layer OCR Perceptual Loss}
In additional, we found that the conditional guidance (such as font or canny hint) added through ControlNet can only control text with relatively large font sizes, while its ability to control smaller fonts is insufficient. The main reason is that during the diffusion process, the model's ability to maintain the font characters within a limited pixel area is inadequate.
Therefore, we introduced multi-layer OCR-aware loss in JoyType. This leverages OCR's strong ability to recognize character shapes, thereby enhancing the diffusion model's ability to maintain the integrity of smaller fonts.
This differs from AnyText~\cite{tuo2023anytext}, which uses OCR loss to improve the recognizability of generated text. JoyType's focus is on maintaining control over smaller fonts, ensuring that the generated text style is sufficiently consistent with the hint conditions.
Therefore, we first map the latent space features on the pixel space to obtain the predicted image ($\hat{x}_{0}$) of the input image ($x_{0}$). Then, using the bounding box (bbox) annotation information given in the training data, we simultaneously crop the text regions of both the ground truth (gt) and the predicted image and input them into the multi-layer OCR-aware module. 
The multi-layer OCR-aware loss $L_{ocr}$ is defined as follows:
\begin{equation} \label{eq::loss-ocr}
    L_{OCR}= \sum_{l}\frac{1}{H_{l}W_{l}}\sum_{h,w}{\left \| \hat{y} _{hw}^{l}-y_{hw}^{l} \right \| }^{2}_{2},
\end{equation}
where $l$ represents the number of convolutional layers in the OCR model. Here, we use the features from the first three convolutional layers of the OCR model, i.e., $l\in[1,3]$. $H_{l}$ and $W_{l}$ denote the height and width of the features in the $l-th$ layer, respectively. For each $(h, w)$ position, we calculate the difference between the predicted and ground truth, and finally take the average.

Overall, the objective function for training JoyType can be formulated as follows:
\begin{equation} \label{eq::loss}
    L = L_{LDM} + \lambda*L_{OCR},
\end{equation}
where $\lambda$ is a weight adjustment parameter used to balance the learning between the latent space and the pixel space. $\lambda$, based on our experiments, is empirically set to $0.1$.

\subsection{Inference Pipeline}
During the inference phase, the image prompt, textual content, and specified areas for text generation are input into the text encoder and Font ControlNet, respectively. The final image is then generated by the VAE decoder. It is particularly worth noting that in terms of text rendering, we do not restrict the content and manner of users' input. Users can specify different languages and any font styles, whether they are common characters or rare ones. We use the DDIM~\cite{songdenoising} sampler with 20 sampling steps.

\begin{table*}[t]
    \centering
    \caption{Performance on easily recognized fonts.}
    \label{tab::font1}
    \setlength{\tabcolsep}{1.3mm}{
        \begin{tabular}{|c|c|c|c|c|c|c|c|c|c|c|c|c|c|c|}
        \hline
         {Fonts} & \multicolumn{2}{c|}{ JDLangZhengTi } & \multicolumn{2}{c|}{ Arial Unicode } & \multicolumn{2}{c|}{ \makecell{SiYuanHeiTi \\ Regular} }  & \multicolumn{2}{c|}{ \makecell{JingNanMaiYuan \\ Ti} }  & \multicolumn{2}{c|}{ \makecell{SiYuanSongTi \\ Regular} }  & \multicolumn{2}{c|}{ \makecell{HuaWenXinWei \\ Ti} } \\
        \hline
         {Metric}                   &   ACC   &    NED      &  ACC     &     NED    &    ACC   &    NED     &   ACC     &    NED     &    ACC    &    NED   &   ACC  &    NED \\
        \hline
          \makecell{Typographic 
          \\ Image}     & 0.8480  &    0.9023   &  0.8494  &   0.9030   &  0.8523  &    0.9039  &   0.8164   &    0.8962  &    0.8489  &  0.9029 & 0.8437 & 0.8998 \\
          \hline
          \makecell{Generated 
          \\ image} & 0.7934  &    0.8772   &  0.7916  &   0.8791   &  0.8054  &    0.8832  &   0.7746   &    0.8764  &    0.7917  &  0.8763 & 0.7569 & 0.8645 \\
        \hline
        \end{tabular}}
\end{table*}

\begin{table*}[t]
    \centering
    \caption{Performance on less recognizable artistic fonts.}
    \label{tab::font2}
    \setlength{\tabcolsep}{1.3mm}{
        \begin{tabular}{|c|c|c|c|c|c|c|c|c|c|c|c|c|c|c|}
        \hline
         {Fonts} & \multicolumn{2}{c|}{ \makecell{ZiXiaoHunMeng\\QuMoLiTi}  } & \multicolumn{2}{c|}{ \makecell{ZiXiaoHunAKai\\TongManTi} } & \multicolumn{2}{c|}{ \makecell{ZiHunGongFuTi} }  & \multicolumn{2}{c|}{ \makecell{ZiHunXiaoMoLi} }  & \multicolumn{2}{c|}{ \makecell{ZiHunBaRan\\ShouShuTi} }  & \multicolumn{2}{c|}{ \makecell{BaShuMoJiTi} } \\
        \hline
         {Metric}                      &   ACC   &    NED      &    ACC     &     NED    &    ACC   &    NED     &    ACC     &    NED       &      ACC     &     NED    &   ACC   &    NED  \\
        \hline
          \makecell{Typographic 
          \\ Image}                    & 0.4508  &    0.7588   &  0.6482    &   0.8449   &  0.6482  &    0.8449  &   0.5664   &    0.8129  &    0.7400  &  0.8705  & 0.3709 & 0.7050 \\
          \hline
          \makecell{Generated 
          \\ image}                    & 0.3168  &    0.6528   &  0.6061    &   0.8067   &  0.4892  &    0.7536  &   0.4103   &    0.7115  &    0.5160  &  0.7641 & 0.2958 & 0.6299 \\
        \hline
        \end{tabular}}
\end{table*}
\section{Experiments}
\label{sec::Experiments}

\subsection{Data Collection}
There is currently a lack of publicly available datasets that exactly tailored our training task, so we built an open source dataset, JoyType-1M. The image in the dataset was sampled from CapOnImage~\cite{gao2022caponimage} and LAION-Glyph-10M~\cite{yang2024glyphcontrol}, which included various images with text, such as street view, natural scenery, and commodity advertisements. We use a Vision Language Model (VLM) CogVLM~\cite{wang2023cogvlm} to regenerate the annotation of each image to align the description of different images. Furthermore, in order to obtain the hint corresponding to each image, we crop out the text areas in the image according to the bounding box information to acquire different text boxes. The canny operator is utilized to extract the edge information in the text boxes respectively, and paste the text boxes back onto the black graph with the same size as the original image to obtain the canny hint. Simultaneously, through using Hi-SAM~\cite{ye2024hi}, which is a unified hierarchical text segmentation model, to process the image, the font hint could be generated. We obtained 1M images in total, the ratio of images from CapOnImage and LAION-Glyph-10M is about 3:1.

\subsection{Implementation Details}
The training framework follows the ControlNet approach, with the model's weights initialized from Stable Diffusion-v1.5. JoyType was trained on the JoyType-1M dataset for 6 epochs using 4 Tesla A100 GPUs. For the ablation experiments, we used JoyType-100K, which is a subset of 100K image-text pairs extracted from JoyType-1M. The image dimensions are set to 
$512\times512$. The AdamW optimizer is used with a learning rate of $1\mathrm{e}{-4}$ and a batch size of 8.

\begin{figure*}[t]
    \centering
    \includegraphics[width=0.98\textwidth]{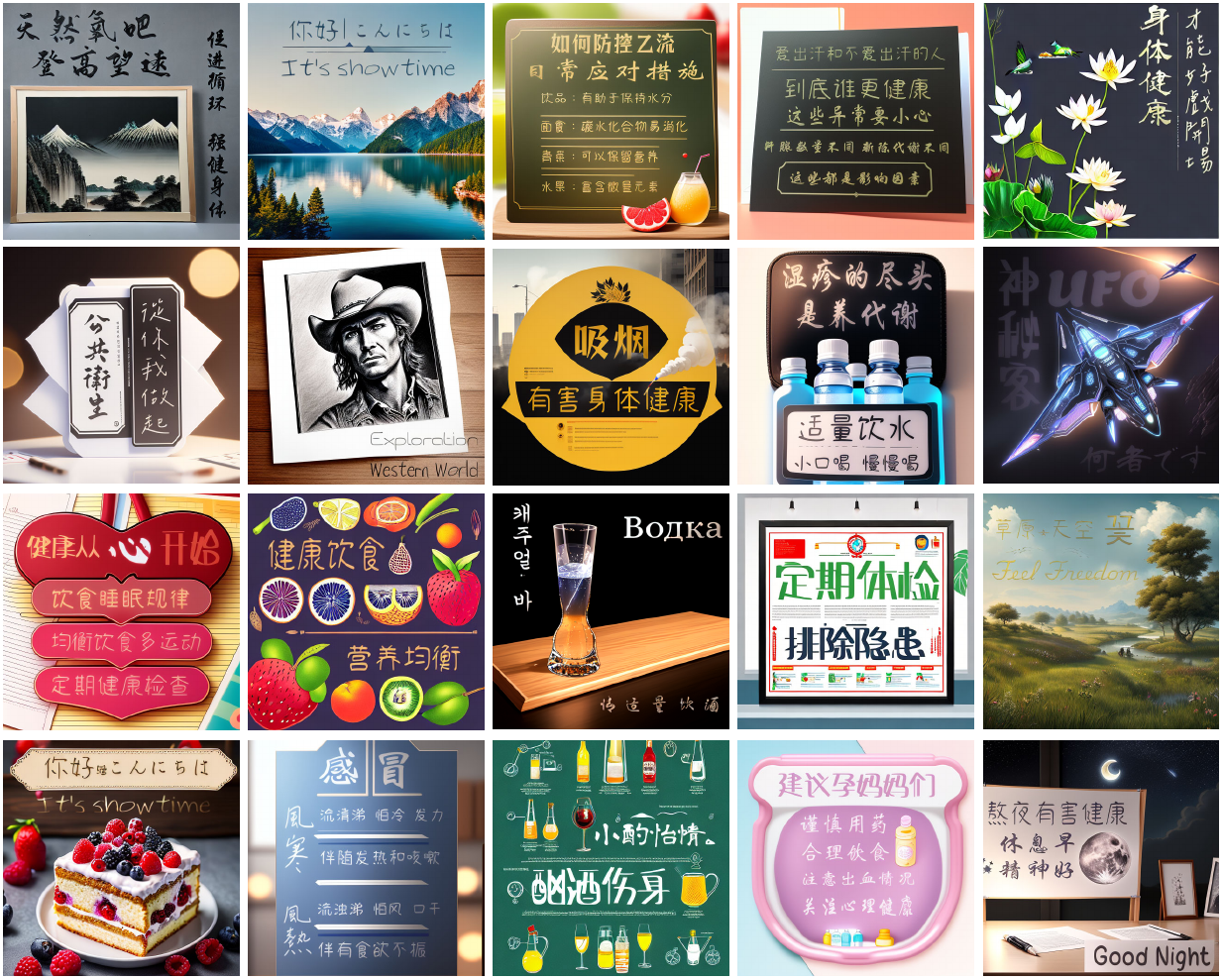} 
    \caption{More examples of JoyType in text generation.}
    \label{fig::moreres}
    \vspace{-0.25cm}
\end{figure*}

\subsection{Baselines and Evaluations}
Focusing on visual text creation, we adopt four popular competing methods including ControlNet~\cite{zhang2023adding}, TextDiffuser~\cite{chen2023textdiffuser2}, GlyphControl~\cite{yang2024glyphcontrol}, and AnyText~\cite{tuo2023anytext}. To ensure fairness in comparison, we use the AnyText-benchmark as the standard evaluation dataset, which using the same positive and negative prompt words. AnyText-benchmark consists of two sub-evaluation sets: wukong and laion. Each set contains 1K test images and is used to evaluate the model's generation capabilities in Chinese and English, respectively. 

For each test, we generate 4 images in a batch. Sentence Accuracy (Acc) and the Normalized Edit Distance (NED) are used as evaluation metrics to assess the recognizability of characters within an image. In this recognition process, we uniformly employ an open-source OCR model~\cite{DuguangOcr}. When evaluating the model's ability to retain fonts, we used 16 different fonts covering Chinese, English, Korean, Japanese, and Russian languages. Specifically for Chinese, we tested 12 font styles, including JDLangZhengTi, Arial Unicode, etc.

\subsection{Font Style Preserving}
Table~\ref{tab::font1} and Table~\ref{tab::font2} demonstrate the ability of our JoyType to maintain font styles. Specifically, we employed over 10 different fonts as hint to create images. 
Typographic Image refers to images printed with selected fonts on a white background, while Generated Image represents images generated by our JoyType. 
To ensure a fair evaluation, all images use the same font size. The closer the performance metrics on the Generated Image are to those on the Typographic Image, the stronger the model's ability to preserve the font style. In Table~\ref{tab::font1}, we used fonts that are generally easily recognizable (e.g., Arial Unicode), while in Table~\ref{tab::font2}, we used less recognizable artistic fonts (e.g., BaShuMoJiTi). 
Compared to Typographic Images, Generated Images achieve similar performance results across most fonts, indicating that JoyType's generated text has relatively high recognizability. This can be intuitively observed from the similar results in the Fig.~\ref{fig::font}.
As can be seen from the figure, regardless of the font's inherent recognizability, JoyType is able to maintain the glyphs, making the rendered text clear and readable. This is thanks to JoyType's use of font-guided conditions during training, allowing the model to simultaneously learn the edge information of the text and the consistency of the font color.

\begin{table}[t]
    \centering
    \caption{Comparison with the SOTAs on two benchmarks.}
    \label{tab::sotas}
    \setlength{\tabcolsep}{0.8mm}{
        \begin{tabular}{|l|c|c|c|c|c|c|}
        \hline
         \multirow{3}{*}{Methods} &  \multicolumn{6}{c|}{Benchmarks} \\
        \cline{2-7}
         
         & \multicolumn{3}{c|}{ wukong } & \multicolumn{3}{c|}{ laion }\\
        \cline{2-7}
                           &   ACC   &    NED      &    FID   &   ACC   &    NED      &    FID \\
        \hline
          GlyphControl     & 0.0327  &    0.0845      &    34.36  &   0.3710 & 0.6680 & 37.84  \\
          TextDiffuser     & 0.0605  &    0.1262      &    53.39  &   0.5921 & 0.7951 & 41.31  \\
          ControlNet       & 0.3620  &    0.6227      &    41.86  &   0.5837 & 0.8015 & 45.41 \\
          AnyText          & 0.6923  &    0.8396      &    31.58  &   0.7239 & 0.8760 & 33.54  \\
        \hline
        JoyType            & 0.7986  &    0.8824      &    26.75  &   0.7971 & 0.9065 & 46.39 \\
        \hline
        \end{tabular}}
    \vspace{-0.25cm}
\end{table}

\subsection{Comparison JoyType with SOTAs}
We compared JoyType with current state-of-the-art methods. Table.~\ref{tab::sotas} presents a comparison between JoyType and GlyphControl, TextDiffuser, ControlNet, and AnyText in both Chinese and English languages. As shown, JoyType significantly outperforms all competitors, as expected. In Chinese text generation, with a better FID score, JoyType's ACC and NED metrics exceed AnyText by $10.63\%$ and $4.28\%$, respectively.
On the FID metric for the laion dataset, JoyType achieved the worst performance. This is because JoyType-1M primarily contains Chinese text, whereas laion is predominantly in English. The significant difference between the JoyType-1M training dataset and the laion evaluation dataset contributed to this result. However, JoyType achieved the highest performance on the ACC and NED metrics, demonstrating its excellent ability to maintain font styles. Even without training on English data, it still shows a strong capability to handle English text.

\subsection{Ablation Studies}
We also verify the impact of different modules on our JoyType’s performance. In Table~\ref{tab::ablation}, JoyType (using hint\_canny) indicates the version of JoyType that uses the Canny hint image.
Two variants are designed as baselines of our JoyType networks: (a) JoyType\_w\_$cogvlm$ is built by replacing the raw short prompt with CogVLM; (b) JoyType\_w\_${hint}_{font}$ is built by replacing glyph hint with font hint. Table~\ref{tab::ablation} shows the comparison results rendering Chinese on wukong benchmark. It can be seen that using the VLM model to rewrite the prompts for images can significantly improve the quality of image generation, with the FID score dropping from 34.00 to 26.75. This improvement is mainly attributed to VLM's ability to provide more detailed image descriptions. 
Compared to JoyType (using hint\_canny), JoyType\_w\_${hint}_{font}$  shows a decrease in ACC, NED, and FID. This is because using the font as a hint tends to generate text with a stroke-like artistic effect around the edges, increasing the diversity of the generated text. Consequently, this leads to a certain degree of decreased recognizability.

\begin{table}[h]
\centering
    \caption{Ablation Studies of JoyType on JoyWords100K. Effectiveness Illustration of each submodule in JoyType.}
    \label{tab::ablation}
    \setlength{\tabcolsep}{1.3mm}{
        \begin{tabular}{|l|c|c|c|}
        \hline
         \multirow{2}{*}{\diagbox{Methods}{Benchmarks}} & \multicolumn{3}{c|}{ wukong }\\
        \cline{2-4}
                           &   ACC   &    NED      &    FID   \\
        \hline
        \makecell{JoyType \\ (using ${hint}_{canny}$)}  & 0.7916  &    0.8791      &    34.00   \\
        \hline
        JoyType\_w\_$cogvlm$    & 0.7986  &    0.8824    &    26.75  \\
        JoyType\_w\_${hint}_{font}$    & 0.7296  &    0.8498    &    31.26   \\
        \hline
        \end{tabular}}
    \vspace{-0.25cm}
\end{table}

\subsection{Evaluation on Small Text Generation.}
\begin{figure}[t]
\centering
\includegraphics[width=0.98\columnwidth]{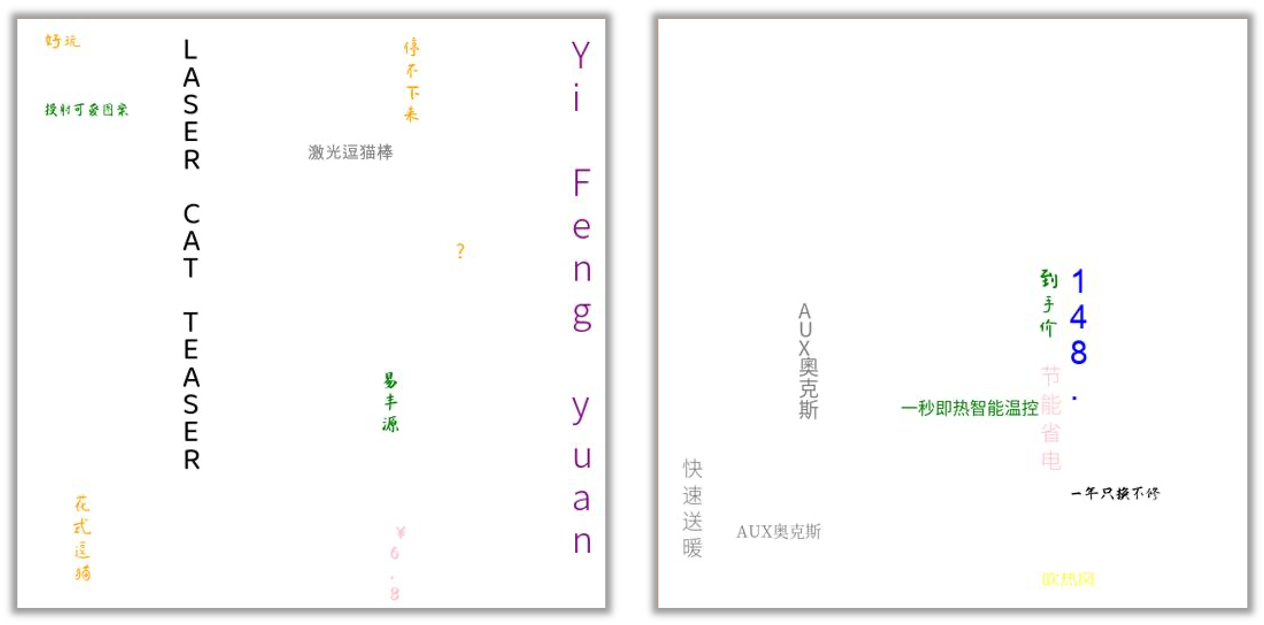} 
\caption{Tiny1K, a manually created small benchmark was established to evaluate the multi-layer OCR-aware module.}
\label{fig::tiny-font}
\end{figure}

Further evaluate JoyType's ability to generate image with smaller font size. To ensure the validity of the evaluation, the generated image resolution is also $512\times512$. However, the different experimental setup involves the font size distribution in the hint images being primarily small fonts. To better evaluate this, we manually constructed an evaluation set Tiny1K specifically for assessing small text generation capability. It includes 1000 white background images of $512\times512$, in which each image contains up to 20 lines of text, with each character being less than 64 pixels in size. Two examples are provided in the image in Fig.~\ref{fig::tiny-font}. The quantitative evaluation results are shown in Table 5.``Typographic Image" indicates the OCR model's evaluation of the text in Tiny1K. Compared to AnyText-benchmark, Tiny1K only contains small fonts, making it more challenging to recognize and better suited to evaluate the model's ability to control small text. JoyType\_wo\_{ocr} represents the JoyType model without the multi-layer OCR-aware loss. Compared to JoyType, not using the OCR-aware loss results in a 1.74\% decrease in ACC and a 1.6\% decrease in NED. 

\begin{table}[t]
\centering
    \caption{Evaluation of JoyType's ability to generate smaller font text.}
    \label{tab::smalltext}
    \setlength{\tabcolsep}{1.3mm}{
        \begin{tabular}{|l|c|c|}
        \hline
         \multirow{2}{*}{\diagbox{Methods}{Benchmarks}} & \multicolumn{2}{c|}{ Tiny1K }\\
        \cline{2-3}
                                &   ACC   &     NED         \\   
        \hline
        Typographic Image       & 0.4850  &    0.5084       \\           
        \hline
        JoyType                 & 0.3013  &    0.4098       \\
        \hline
        JoyType\_wo\_${ocr}$    & 0.2839  &    0.3938       \\
        \hline
        \end{tabular}}
    \vspace{-0.25cm}
\end{table}

\subsection{Discussion and Limitations}
Our advantage lies in maintaining the font styles in text rendering. Therefore, JoyType accepts instructions in any font style and preserves the font style during the image generation process. This is entirely different from previous methods, such as Glyphcontrol and AnyText, which use default font styles and generate uncontrollable font styles. 
Due to JoyType's adoption of a ControlNet-like model design, it is endowed with excellent scalability. This allows it to be compatible with various open-source diffusion models, facilitating the generation of images in a multitude of styles. More examples of JoyType in text generation are shown in Fig.~\ref{fig::moreres}
Therefore, to some extent, compared to previous methods, JoyType represents a more practical and scalable approach, making it easier to integrate into actual workflows.
In this work, JoyType is trained based on the Stable Diffusion v1.5 model. To further enhance the quality of generated images, a feasible approach is to use more advanced base models, such as SDXL or models based on the DiT architecture, to render text on higher resolution images. This approach can reduce the difficulty of maintaining glyphs for small text to some extent. However, since this falls outside the scope of model learning, it is not discussed in this paper.
In the field of text rendering, another direction is intelligent text layout. A common approach relies on LLM models to output the position for the rendered text. However, since the core content is about how to finetuning a LLM, it is also not within the scope of our discussion in this work.

\section{Conclusion}
\bigskip
\noindent 

This paper presents a novel multilingual visual text creation method, dubbed JoyType, which aims to generate images effectively rendering readable texts.
First, we introduce a novel architecture, termed Font ControlNet, designed to maintain font style. This innovation enhances the diffusion model's capability to preserve text font styles across various languages, multiple font styles, and a spectrum of character frequencies, from common to rare characters. To train JoyType, we have compiled a new dataset, JoyType-1M, which comprises 1 million pairs of text-image-hint representations.
Additionally, a multi-layer OCR perceptual loss is introduced into JoyType to bolster the model's proficiency in rendering text with small-sized fonts.
Finally, the effectiveness of JoyType is well demonstrated by comprehensive experiments conducted on different benchmarks, showcasing its superior performance in generating accurate and visually appealing text across diverse languages and font styles.

%

\bibliography{aaai25}

\end{document}